\documentclass[11pt]{article}

\usepackage[final]{acl}

\usepackage{times}
\usepackage{latexsym}
\usepackage{amsmath}
\usepackage{booktabs}
\usepackage{amssymb}
\usepackage[T1]{fontenc}

\usepackage[utf8]{inputenc}

\usepackage{microtype}

\usepackage{inconsolata}

\usepackage{graphicx}

\title{What Survives Distillation? A Mechanistic Interpretability Study of GPT-2 and DistilGPT-2}

\author{Nabeelah Faumi  \\
  Informatics Institute of Technology\\
57, Ramakrishna Road, Colombo 06,\\
Sri Lanka\\
  \texttt{nabeelahfaumi@gmail.com } \\\And
  Deshan Sumanathilaka \\
  Informatics Institute of Technology\\
  57, Ramakrishna Road, Colombo 06,\\
 Sri Lanka\\
  \texttt{deshan.s@iit.ac.lk} \\}

\begin{document}
\maketitle
\begin{abstract}
Knowledge distillation compresses large language models into compact students that preserve aggregate metrics, yet the mechanistic consequences of depth-reduction distillation remain largely uncharacterized. This work applies linear CKA, logit lens analysis, attention pattern decomposition, sparse autoencoder feature statistics, and activation patching to compare GPT-2 and its six-layer student DistilGPT-2 across two canonical circuits: the Indirect Object Identification (IOI) circuit and the Greater-Than (GT) circuit. The results reveal a compression-collapse dichotomy: the IOI circuit suffers complete mechanistic collapse (DistilGPT-2: 35.2\% accuracy vs.\ 94.8\%, zero name-mover preservation), while the GT circuit survives with 87.7\% task performance recovery through active remapping of the teacher's layer-9 comparison head into student layer~4 (mean Pearson $r = 0.945$). These findings demonstrate that circuit compressibility governs mechanistic survival under distillation, and that aggregate metrics are insufficient diagnostics for circuit integrity.

\end{abstract}

\section{Introduction}
\label{sec:intro}
Large language models (LLMs) exhibit remarkable capabilities, but their sheer scale poses significant computational challenges for deployment. Knowledge distillation trains a compact student on soft targets from a larger teacher, enabling compression while mimicking the teacher's generalization \citep{75}. While distillation often successfully preserves aggregate metrics like validation perplexity, it remains an open question how severe architectural truncation affects the precise, underlying algorithms the model has learned. Prior work establishes that compression can degrade circuit integrity and alter feature representations \citep{86, 11}, but the specific mechanistic consequences of distillation-induced depth reduction remain largely uncharacterized. Mechanistic interpretability provides a framework for addressing this blind spot by reverse-engineering neural networks into human-understandable computational subgraphs, known as circuits \citep{92, 14}.

This work investigates the mechanistic impact of depth-reduction distillation. It focuses on DistilGPT-2 \citep{102}, a 6-layer student distilled from the 12-layer GPT-2 teacher \citep{106}. This drastic depth reduction poses a critical structural challenge: many well-documented higher-order reasoning behaviors in GPT-2 depend crucially on late-layer interactions. For example, both the Indirect Object Identification (IOI) circuit \citep{94} and the Greater-Than (GT) circuit \citep{95} rely heavily on highly specialized attention heads located in layers 9-11. Because the student model is only 6 layers deep, these components have no direct structural counterpart. This raises two central research questions: do the internal representations of DistilGPT-2 remain similar to those of GPT-2 after depth halving (RQ1), and does distillation preserve the computational circuits responsible for higher-order reasoning, and if so, how does the student compensate for the absence of essential upper-layer components (RQ2).

To answer this, a suite of representational and causal probes, including representational similarity analysis, direct logit attribution \citep{56}, and activation patching \citep{21}, is applied to track circuit preservation across the distillation gap. A natural question is whether circuit preservation matters at all if task performance is maintained. We argue it does, for two reasons. First, the IOI and GT circuits are among the most causally validated subgraphs in any language model, established through convergent evidence from activation patching, path patching, and direct logit attribution across independent studies \citep{94,95}; they are not arbitrary correlational subgraphs but confirmed computational mechanisms. Second, and more critically, the IOI result we report is not a performance degradation but a sign reversal: DistilGPT-2 achieves a negative mean logit difference ($-0.467$), meaning the model is actively misled rather than merely uncertain,  failure mode entirely invisible to aggregate metrics, the core motivation for circuit-level analysis.

We find a striking dichotomy in how different computational mechanisms survive depth halving. One circuit collapses entirely, the other is actively remapped into the student's shallower stack, detailed in \S\ref{sec:results} and \S\ref{sec:discussion}. By contrasting a circuit that collapses with one that compresses, this sheds light on the inductive biases of the distillation process. 

The main contributions are as follows:
\begin{itemize}
    \item Provide a comprehensive, circuit-level analysis comparing a depth-distilled model to its teacher, quantifying the mechanistic impact of structural truncation.
    \item Demonstrate that complex, multi-stage computational pipelines (such as the IOI circuit) fail to survive standard distillation when critical late-layer components are excised.
    \item Discover that the student model can successfully compress and remap specific cognitive algorithms (such as the GT circuit), migrating late-layer mechanisms into its available upper layers to preserve task functionality.
\end{itemize}

In the following sections, we review related work, outline our experimental setup, present our results, and conclude with a discussion and conclusion.

\section{Related Work}
\textbf{Mechanistic Interpretability of GPT-2.} Circuit analysis decomposes neural networks into functionally distinct sub-networks using causal interventions \citep{14, 13}. In GPT-2, two circuits are directly relevant to this work: the Indirect Object Identification (IOI) circuit \citep{94}, a multi-stage pipeline of name-mover, S-inhibition, and induction heads spanning layers 0-11; and the Greater-Than (GT) circuit \citep{95}, in which year-encoding heads feed comparison heads in layers 9-11. Activation patching \citep{21}, and path patching \citep{25} establish causal responsibility at the head and pathway level, respectively, while Direct Logit Attribution \citep{56} provides a complementary non-interventional decomposition. Sparse autoencoders \citep{57, 112} extend circuit analysis to the feature level by decomposing polysemantic residual-stream activations into monosemantic feature dictionaries.

\textbf{Knowledge Distillation.} Knowledge distillation trains a compact student on soft targets from a larger teacher, enabling it to mimic the teacher's generalization while maintaining a smaller architecture \citep{75}. DistilGPT-2 \citep{102} reduces GPT-2's depth to six layers by initializing from every other teacher layer and combining a language-modeling and distillation objective, thereby preserving embedding dimensionality throughout. 

\textbf{Compression and Interpretability.} Prior work establishes that compression can degrade circuit integrity and alter feature representations \citep{86,11}, but focuses primarily on pruning and quantization \citep{80,81}. SAE-based analysis reveals representational changes under compression can be invisible to standard metrics \citep{84,85}. Most closely related, \citet{114,115,116} show distillation induces simplicity bias, component reorganisation, and unique feature directions across model families. 

\section{Background and Models}
\label{sec:background}
\textbf{Mechanistic interpretability} \citep{92,14} reverse-engineers neural network algorithms at the level of attention heads and residual-stream components. The \emph{residual stream} \citep{92} treats activations as a running sum of additive contributions from each head and MLP, enabling direct logit attribution (DLA). A \emph{circuit} \citep{14} is a minimal subgraph causally sufficient for a given behaviour. The IOI and GT circuits both depend critically on upper-layer interactions structurally absent in a six-layer student.


\textbf{Representational similarity.} Debiased linear CKA \citep{100} compares activations across models. Let
$H_X=\widehat{\mathrm{HSIC}}(XX^\top,XX^\top)$,
$H_Y=\widehat{\mathrm{HSIC}}(YY^\top,YY^\top)$, and
$H_{XY}=\widehat{\mathrm{HSIC}}(XX^\top,YY^\top)$.
Then,
\begin{equation}
\mathrm{CKA}(X,Y)=\frac{H_{XY}}{\sqrt{H_X H_Y}}.
\end{equation}
Here, $\widehat{\mathrm{HSIC}}$ denotes the unbiased Hilbert--Schmidt Independence Criterion estimator \citep{111}, which is invariant to orthogonal transformations and isotropic scaling.

\textbf{Sparse autoencoders} \citep{112,57} decompose residual-stream activations into sparse, interpretable feature directions. Pre-trained GPT-2 SAEs \citep{20} (\texttt{gpt2-small-res-jb}) are applied to both models at matched layers as a fixed reference basis, testing whether DistilGPT-2's representations lie in the same sparse feature subspace.

\textbf{Models.} GPT-2 \citep{106} is a 12-layer decoder-only transformer ($d_\text{model}{=}768$, 12 heads, ${\sim}117$M parameters) and serves as the \emph{teacher}. DistilGPT-2 \citep{102}, the \emph{student}, retains the same embedding dimensionality while halving the depth to 6 layers (${\sim}82$M parameters), initialised from every other teacher layer and fine-tuned with a combined language-modelling and soft-label distillation objective. IOI circuit heads in layers 6--11 and GT circuit heads in layers 9--11 have no direct counterpart in the student.

\section{Methods}
\label{sec:methods}
To answer RQ1, debiased linear CKA (\S\ref{sec:cka}), logit lens analysis (\S\ref{sec:logitlens}), and attention pattern analysis (\S\ref{sec:attention}) are applied. To answer RQ2, circuit-level analysis of the IOI and GT tasks is conducted via activation patching and direct logit attribution (\S\ref{sec:methods-ioi}-\ref{sec:gt_methods}). All experiments use $n{=}1{,}000$ samples from the TinyStories validation set \citep{91}, filtered to $[64, 256]$ tokens, processed via TransformerLens \citep{108} on a single NVIDIA L4 GPU.

\subsection{Representational Similarity via CKA}
\label{sec:cka}
Debiased linear CKA \citep{100} is computed between residual-stream activations at matched layers. Activations are mean-pooled over non-padding positions:
\begin{equation}
    \mathbf{h}_\ell^{(i)} = \frac{\sum_{t=1}^{T_i} \mathbf{m}_t \odot \mathbf{h}_{\ell,t}^{(i)}}{\sum_{t=1}^{T_i} \mathbf{m}_t},
\end{equation}
yielding $X_\ell, Y_\ell \in \mathbb{R}^{n \times 768}$ in \texttt{float64}. CKA is computed via the unbiased HSIC estimator \citep{111} with variance terms clamped to $10^{-16}$.

\textbf{Layer correspondence.} Two schemes are used: \textbf{same-index} compares layer $\ell$ of each model for $\ell \in \{0,3,5\}$; \textbf{proportional} maps DistilGPT-2 layer $d$ to GPT-2 layer $g = \lfloor d \cdot 11/5 \rceil$, giving pairs (G0/D0), (G2/D1), (G4/D2), (G7/D3), (G9/D4), (G11/D5). 

\subsection{Logit Lens Analysis}
\label{sec:logitlens}
The tuned logit lens \citep{97,110} projects each intermediate hidden state through the final layer norm and unembedding matrix:
\begin{equation}
    \hat{p}_{\ell,t}^{(i)} = \mathrm{softmax}\!\left(W_U\, \mathrm{LN}_f\!\left(
    \mathbf{h}_{\ell,t}^{(i)}\right)\right).
\end{equation}
Next-token prediction accuracy $\mathrm{Acc}_\ell$ and relative accuracy $r_\ell = \mathrm{Acc}_\ell^\text{distil} / \mathrm{Acc}_\ell^\text{GPT-2}$ are computed at each layer under both mapping schemes.

\subsection{Attention Pattern Analysis}
\label{sec:attention}
Per-head attention patterns are extracted via TransformerLens hooks. Three metric families are computed: Shannon entropy $H_{l,h}$ and top-$k$ attention mass to quantify head collapse; mean BOS, previous-token, and EOS weights to characterise structural head types \citep{92}; and per-sample cosine similarity between flattened $T{\times}T$ attention matrices at matched layer pairs, alongside per-head entropy change $\Delta H_{l,h} = H_{d,h}^{\mathrm{distil}} - H_{l,h}^{\mathrm{GPT2}}$.

\subsection{SAE-Based Feature Analysis}
\label{sec:sae}
Pre-trained GPT-2 SAEs (\texttt{gpt2-small-res-jb}, \citealt{20,96}) are applied to both models' activations at matched layers as a fixed reference basis, testing whether DistilGPT-2's representations lie in the same sparse feature subspace as GPT-2. Four metrics are computed: \textit{activation rate} $r_f$; \textit{L0 sparsity} $\bar{L}_0$ and change $\Delta L_0$; \textit{feature survival} (Pearson correlation per feature, excluding $\sigma 10^{-8}$); and \textit{magnitude drift} $\delta_f = \bar{z}_f^{\mathrm{distil}} / (\bar{z}_f^{\mathrm{GPT2}} + \varepsilon)$.

\subsection{Indirect Object Identification Circuit Analysis}
\label{sec:methods-ioi}
\label{tab:IOIheads}
\begin{table*}[h]
\centering
\small
\caption{IOI circuit heads by functional class and availability in DistilGPT-2 (layers 0-5 only).}
\begin{tabular}{lll}
\toprule
Class & GPT-2 heads & Available in DistilGPT-2 \\
\midrule
Previous token     & (0,1),(2,2),(4,11) & All 3 \\
Duplicate token    & (0,10),(3,0)       & All 2 \\
Induction          & (5,1),(5,5),(6,9),(7,2),(7,10) & 2 of 5 \\
S-inhibition       & (7,3),(7,9),(8,6),(8,10)       & None \\
Name mover         & (9,6),(9,9),(10,0),(10,6),(11,2) & None \\
Neg.\ name mover   & (10,7),(11,10)                  & None \\
Backup name mover  & (9,0),(9,1),(10,2),(10,10),(11,9) & None \\
\bottomrule
\end{tabular}
\end{table*}

The IOI task \citep{94} uses prompts of the form \textit{``[Name$_1$] and [Name$_2$] went to the [place]. [Name] gave a [object] to''} with 1,000 clean and 1,000 corrupted (ABC) prompts from 15 syntactic templates; the primary metric is logit difference $\Delta(x) = \ell_{\mathrm{IO}}(x,t^*) - \ell_{\mathrm{S}}(x,t^*)$. Of the 24 IOI circuit heads, 19 (all in layers 6--11) are structurally absent in DistilGPT-2; analysis is restricted to the 7 heads in layers 0--5 and 3 inter-head paths whose endpoints lie within layers 0--5. Each head's causal contribution is measured by replacing its corrupted-run output with the clean-run activation \citep{21,94}; inter-head path patching \citep{25} injects the clean-corrupt difference at the receiver's residual stream. Significance is assessed with paired $t$-tests and Wilcoxon tests (Bonferroni correction); effect sizes reported as Cohen's $d$. DLA contributions $\delta^{l,h}_{\mathrm{IOI}} = (\mathbf{W}_U \mathrm{LN}_f(\mathbf{v}^{l,h}))_{\mathrm{IO}} - (\mathbf{W}_U \mathrm{LN}_f(\mathbf{v}^{l,h}))_{\mathrm{S}}$ are computed via a single \texttt{einsum} over cached head outputs. To test whether any DistilGPT-2 head assumes the name-mover role, CKA and Pearson correlation between each GPT-2 name-mover head's attention patterns and the top-20 DistilGPT-2 components (by DLA) are computed on a discovery set (500 prompts) and confirmed on a held-out evaluation set (500 prompts); the name-mover score $\mathrm{NM}(l,h) = \mathbb{E}[\sum_{i \in \mathrm{IO}} \alpha^{l,h}_{t^*,i} - \sum_{j \in \mathrm{S}} \alpha^{l,h}_{t^*,j}]$ quantifies IO-selective attention.

\subsection{Greater-Than Circuit Analysis}
\label{sec:gt_methods}
The GT task \citep{95} uses prompts of the form \textit{``The [noun] lasted from the year [XX][YY] to the year [XX]''} with 1,000 clean and 1,000 corrupted prompts; the primary metric is GT probability mass $\mathrm{GTProb}(x,t^*) = \sum_{yy > YY} p(\texttt{yy} \mid x, t^*)$. Of the six GT circuit heads, only the two year-encoding heads $\{(0,1),(1,5)\}$ exist in DistilGPT-2; all comparison and output-computation heads in layers 9--11 are absent. The same patching procedure as IOI is applied, with scores normalised by the noise ceiling $\hat{s}_{l,h} = s_{l,h} / (\mathrm{GTProb}_{\mathrm{clean}} - \mathrm{GTProb}_{\mathrm{corrupt}})$. GT DLA scores $\delta^{l,h}_{\mathrm{GT}} = \sum_{yy > YY} \mathrm{softmax}(\mathbf{W}_U \mathrm{LN}_f(v^{l,h}))_{yy}$ give each component's contribution to GT probability mass. The same discovery-evaluation matching procedure tests whether any DistilGPT-2 head replicates GPT-2's comparison head (9,1).

\section{Results}
\label{sec:results}
\subsection{Perplexity Baseline}

Table~\ref{tab:perplexity} reports perplexity for both models on TinyStories and WikiText-2, establishing that aggregate metrics appear reasonable despite the circuit-level collapse documented in the main paper.

\begin{table*}[h]
\centering
\small
\caption{Perplexity comparison on TinyStories validation and WikiText-2 test sets.}
\label{tab:perplexity}
\begin{tabular}{lcccc}
\toprule
Model & Layers & Params & PPL (TinyStories) & PPL (WikiText-2) \\
\midrule
GPT-2       & 12 & 124M & 11.54 & 33.02 \\
DistilGPT-2 & 6  & 82M  & 17.55 \scriptsize{(+52.1\%)} 
                         & 49.33 \scriptsize{(+49.4\%)} \\
\bottomrule
\end{tabular}
\end{table*}

The perplexity gap ($\Delta$\,=\,6.01 on TinyStories, $\Delta$\,=\,16.31 on WikiText-2) is substantial in absolute terms but does not signal mechanistic failure: the student retains $87.7\%$ GT circuit performance while showing $52\%$ higher perplexity, and loses the IOI circuit entirely while still producing fluent text. This illustrates the central claim that perplexity is an insufficient diagnostic for circuit integrity.

\subsection{Representational Similarity via CKA}
\label{sec:results_cka}
\begin{figure}[t]
    \centering
    \includegraphics[width=\linewidth]{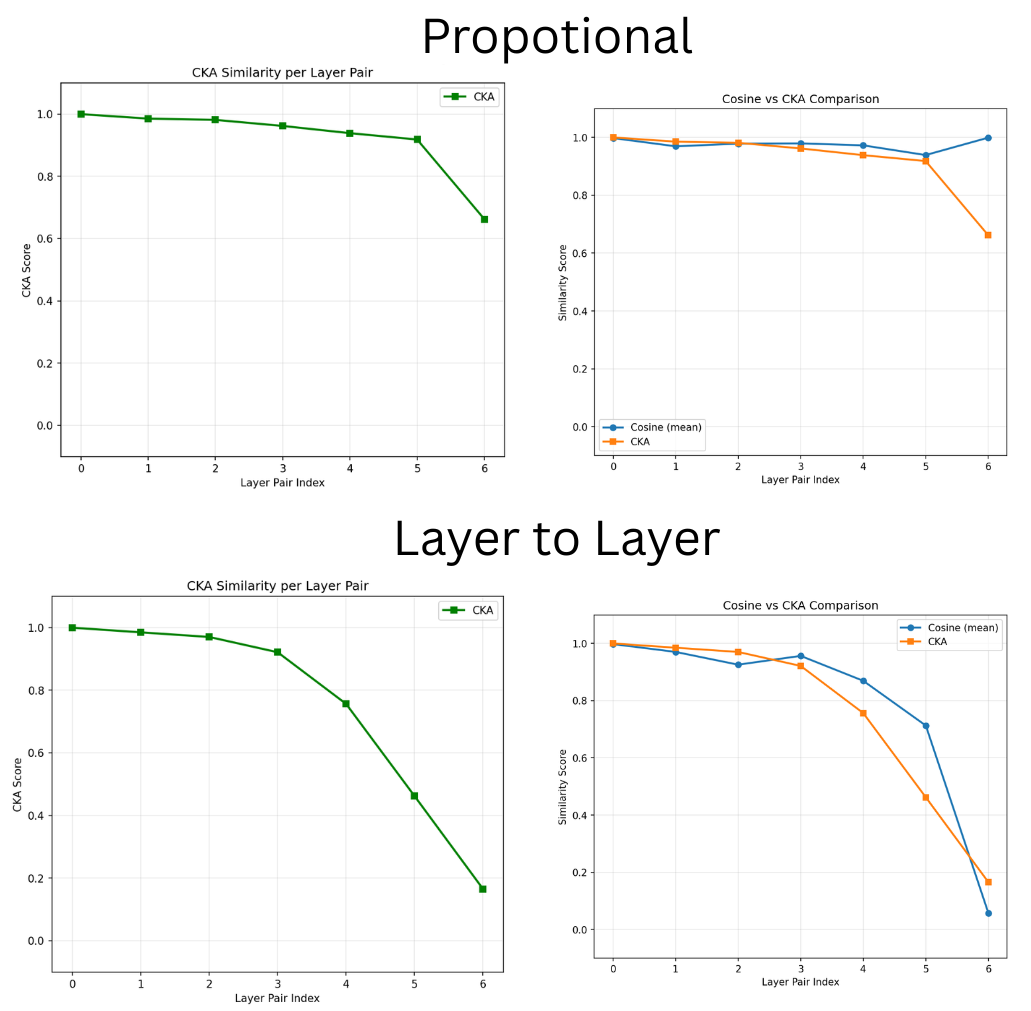}
    \caption{Layer-wise CKA similarity under proportional mapping (top) and same-index mapping (bottom). Late-layer CKA collapses to $0.314$ at layer~5 under same-index matching and is partially recovered under proportional mapping ($0.790$). Cosine similarity results agree in direction throughout.}
    \label{fig:cka_main}
\end{figure}
Under \textbf{proportional mapping} (Figure~\ref{fig:cka_main}), similarity degrades monotonically with depth: early layers are near-perfectly aligned (CKA $=0.9998$, cosine $=0.9977$), middle layers retain strong similarity (CKA $=0.967$), and late layers fall to CKA $=0.790$. Overall means are CKA $=0.921\pm0.109$ and cosine $=0.977\pm0.019$. Under \textbf{same-index mapping}, late-layer degradation is severe (CKA $=0.314$, cosine $=0.385$ at layer 5), with overall means dropping to CKA $=0.752\pm0.298$. The contrast between schemes is informative: the same-index collapse at layer 5 is substantially recovered under proportional mapping (CKA $0.314\to0.790$), showing the student compresses the teacher's representational trajectory into fewer steps rather than truncating it.

\subsection{Logit Lens}
\label{sec:results_logit}
\begin{table*}[h]
\centering
\small
\caption{Extended logit lens accuracy comparison including mean probability assigned to correct next token ($\bar{p}_\ell$).}
\label{tab:logit_proportional}
\begin{tabular}{llrrrrr}
\toprule
Pair & GPT-2 $\ell$ & Distil $\ell$ & $\Delta$ Acc & $r_\ell$ \\
\midrule
D0/G0  & 0  & 0 & $-$0.005 & 1.13\\
D1/G2  & 2  & 1 & $-$0.039 & 2.10\\
D2/G4  & 4  & 2 & $-$0.021 & 1.32\\
D3/G6  & 6  & 3 & $-$0.030 & 1.28\\
D4/G8  & 8  & 4 & $-$0.068 & 1.35\\
D5/G11 & 11 & 5 & $+$0.006 & 0.97\\
\bottomrule
\end{tabular}
\end{table*}
Under proportional mapping, DistilGPT-2 consistently achieves higher relative accuracy than its matched GPT-2 counterpart at every layer except the final (Table~\ref{tab:logit_proportional}), with $r$ ranging from $0.97$ (D5/G11) to $2.10$ (D1/G2). Under same-index mapping, the picture reverses: DistilGPT-2 underperforms at every layer with $r$ up to $3.95$ at L4, reflecting how much representational work the teacher's upper layers perform.

\subsection{Attention Pattern Analysis}
\label{sec:results_attention}
\begin{figure}[t]
    \centering
    \includegraphics[width=\linewidth]{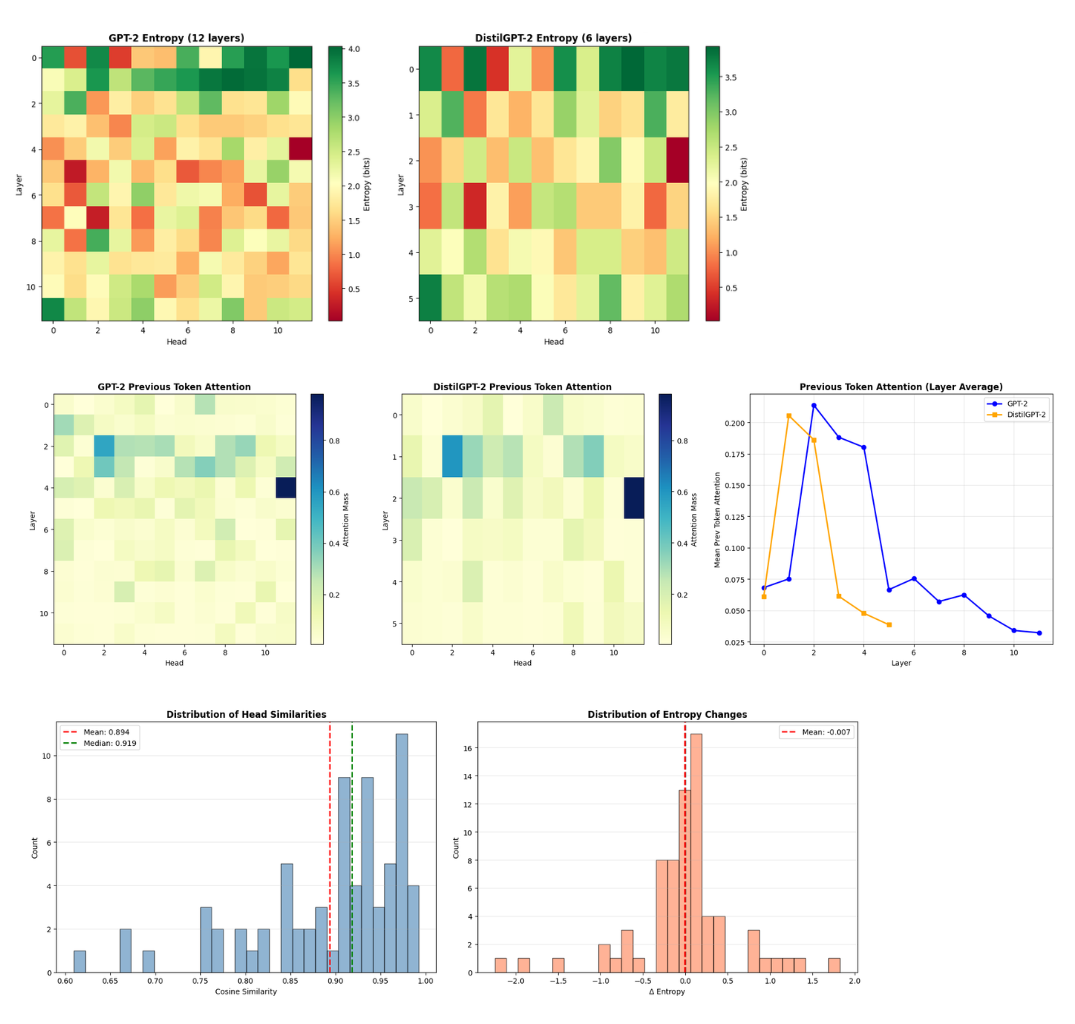}
    \caption{
    \textbf{Attention pattern analysis.}
    \textit{Top}: entropy heatmaps for GPT-2 and DistilGPT-2 (Q1).
    \textit{Middle}: previous-token attention heatmaps and layer averages; student peaks earlier at layer~1 (Q2).
    \textit{Bottom}: distributions of per-head cosine similarity (mean $0.894$) and entropy change (mean $-0.007$ bits) under proportional mapping (Q3).}
    \label{fig:attention_main}
\end{figure}
Both models show broadly similar entropy profiles (GPT-2: $2.001$ bits, DistilGPT-2: $2.094$ bits), with DistilGPT-2 heads marginally \emph{more} distributed than the teacher's. Previous-token attention diverges at layer~1 (student $\approx0.36$ vs teacher $\approx0.24$), suggesting the student front-loads local token-copying into early layers. Under proportional mapping, per-head cosine similarity between matched attention patterns is $0.894 \pm 0.086$ (range $[0.610, 0.993]$); same-index mapping drops to $0.631 \pm 0.255$ (range $[0.038, 0.981]$), consistent with the CKA finding that same-index pairs are poorly matched beyond the first few layers.

\subsection{SAE Feature Analysis}
\label{sec:results_sae}

Under same-index matching, feature survival and cosine similarity degrade sharply with depth (Appendix Table~\ref{tab:sae_summary}): at layer~5, survival is $0.086$ and reconstruction error is $34.9\times$ the teacher's. Under proportional matching, the early pair (G1/D0) shows near-zero survival ($0.0002$), reflecting feature divergence within that specific layer pair rather than a matching scheme artefact; mid- and late-pipeline pairs recover to survival $\approx0.53$ and cosine $0.927$, with magnitude drift $\delta\approx0.51$--$0.55$.

\subsection{Indirect Object Identification Circuit Preservation}
\label{sec:results_ioi}
\begin{figure}[ht]
    \centering
    \includegraphics[width=\linewidth]{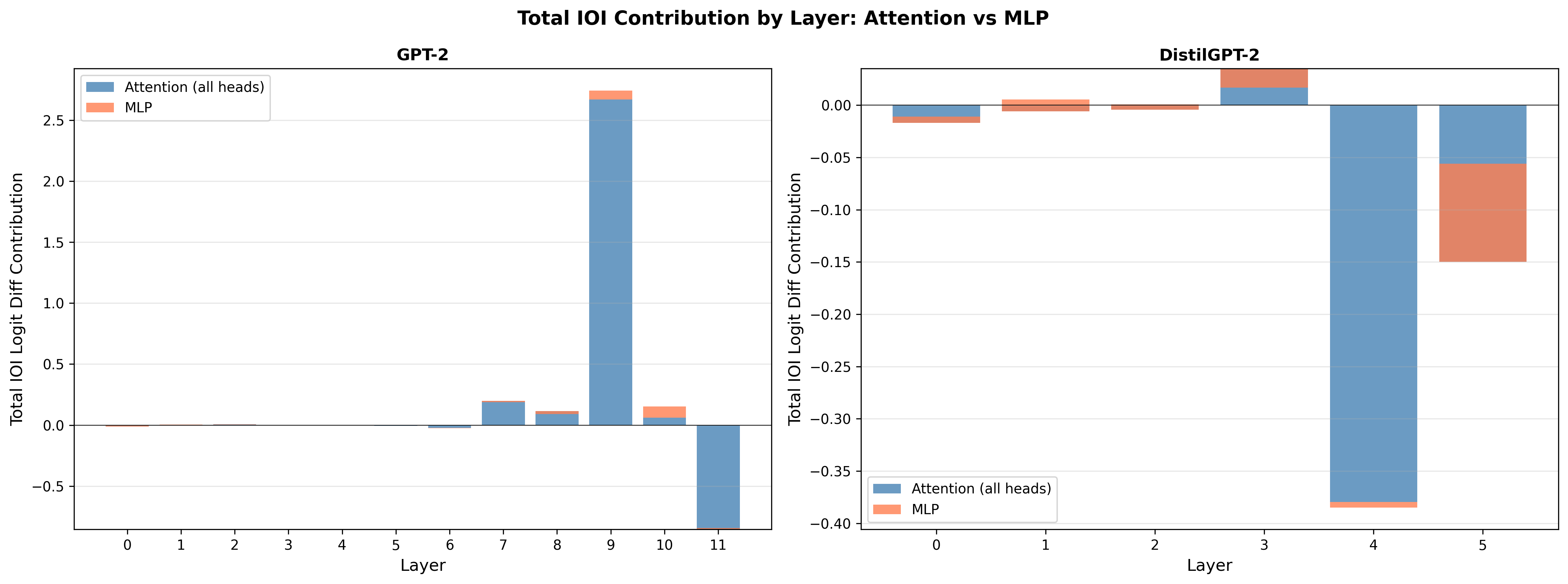}
    \caption{DLA across layers for the IOI task. GPT-2 shows strong positive contributions from layers 9-10; DistilGPT-2 shows near-zero attribution across all available layers.}
    \label{fig:dla_stacked}
\end{figure}

\begin{figure}[ht]
    \centering
    \includegraphics[width=\linewidth]{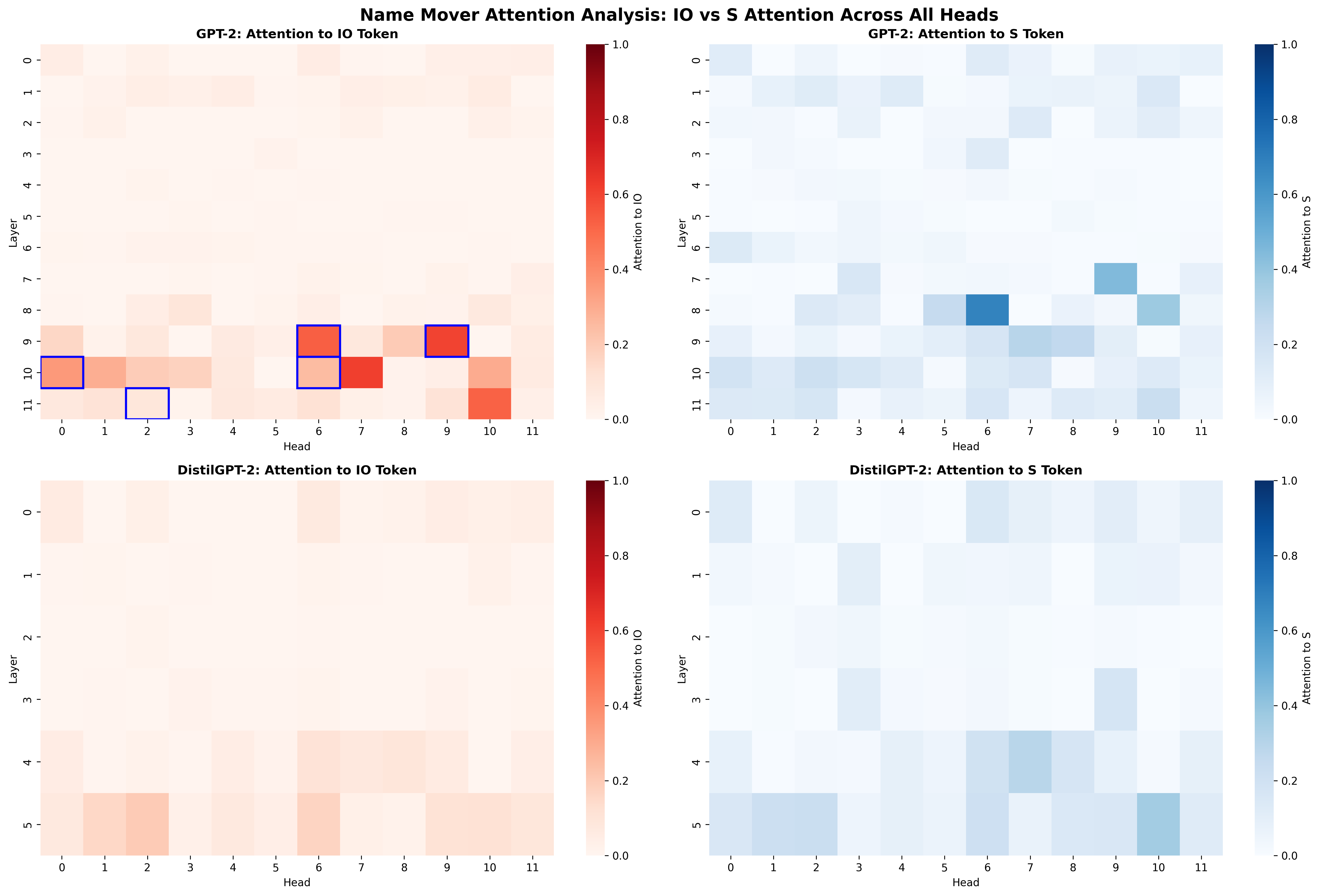}
    \caption{Attention heatmaps for GPT-2 name-mover heads and their closest DistilGPT-2 matches. Matched student heads fail to exhibit IO-selective attention.}
    \label{fig:nm_heatmaps}
\end{figure}

\begin{table*}[h]
\centering
\small
\caption{Full name-mover head matching results across discovery and evaluation sets.}
\label{tab:ioi}
\begin{tabular}{lllrrrr}
\toprule
GPT-2 & Distil & Disc CKA & Eval CKA & Disc $r$ & Eval $r$ & NM Pres\% \\
\midrule
L9H6  & L0H1 & 0.4659 & 0.4592 & $-$0.1535 & $-$0.1514 & 0.0\% \\
L9H9  & L0H1 & 0.3209 & 0.3261 & $-$0.1391 & $-$0.1378 & 0.0\% \\
L10H0 & L1H6 & 0.5077 & 0.5150 & 0.4597  & 0.4417  & $-$14.6\% \\
L10H6 & L1H6 & 0.3629 & 0.3926 & 0.6507  & 0.6484  & $-$21.3\% \\
L11H2 & L1H6 & 0.2073 & 0.1784 & 0.7802  & 0.7781  & nan \\
\bottomrule
\end{tabular}
\end{table*}
GPT-2 achieves a mean logit difference of $2.942\pm1.797$ and $94.8\%$ accuracy on the IOI task. DistilGPT-2 fails completely: logit difference $-0.467\pm1.335$, accuracy $35.2\%$. This is below the $50\%$ chance level for a two-name choice (one-sided binomial test, $n=1{,}000$, $p<0.001$), so the student is systematically wrong rather than guessing. The negative logit difference indicates the model actively favours the subject name over the indirect object, a sign reversal, not merely a performance gap. One possible explanation, which we do not test, is a repeated-name heuristic: in IOI prompts the subject appears twice and the indirect object once, so a model that favours frequently seen names would produce exactly this sign. DLA confirms no component provides meaningful positive attribution the student's peak layer contribution is $0.0168$ vs GPT-2's $2.74$ at layer 9, roughly $160\times$ smaller, and all IOI circuit classes score approximately zero (Figure~\ref{fig:dla_stacked}). Activation patching of the 7 available heads (Table~\ref{tab:ioi}: three previous-token, two duplicate-token, two induction) yields negligible recovery; induction heads show a small effect (Cohen's $d=0.298$, $p<0.001$) but the critical PT$\to$IND pathway is non-significant ($d=0.031$, $p=0.329$). Name-mover matching (Figure~\ref{fig:nm_heatmaps}) identifies DistilGPT-2 candidates with CKA between $0.18$ and $0.52$, and attention-pattern correlation that ranges from weakly negative ($r\approx-0.15$ for the layer-9 heads) to strongly positive ($r\approx0.44$--$0.78$ for the layer-10 and layer-11 heads). However, no matched head reproduces IO-selective attention: name-mover score preservation is at most $0\%$ for every match, so pattern similarity alone does not indicate that the name-mover function has been remapped. This confirms that soft-label distillation does not force remapping of critical late-layer IOI functionality.

\subsection{Greater-Than Circuit Preservation and Compression}
\label{sec:results_gt}
\begin{figure}[ht]
    \centering
    \includegraphics[width=\linewidth]{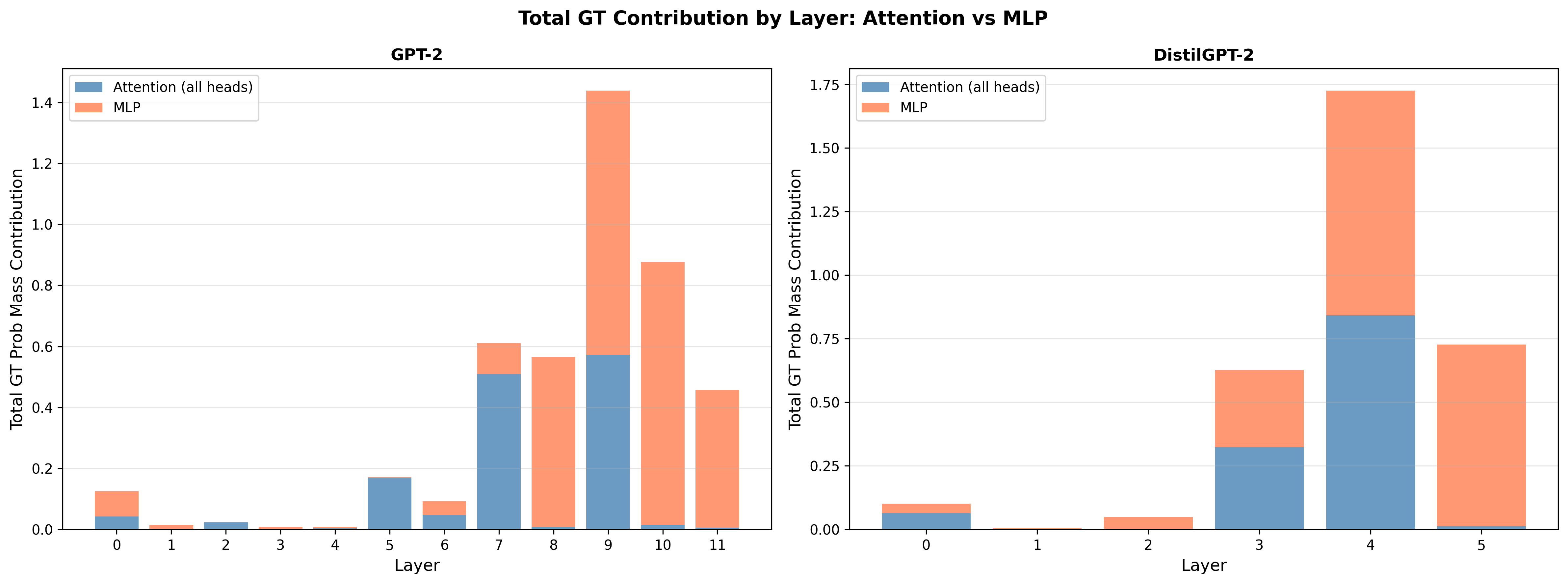}
    \caption{DLA for the GT task. DistilGPT-2 shifts peak attribution to layer 4, functionally replacing GPT-2's layer 9.}
    \label{fig:gt_dla_stacked}
\end{figure}

DistilGPT-2 maintains $87.3\%$ accuracy and GT probability mass $0.750$ (vs GPT-2's $94.0\%$ and $0.855$), recovering $87.7\%$ of teacher performance. DLA reveals an active circuit shift: GPT-2's peak contribution is at layer 9 (L9H1, L9MLP, L10MLP), while DistilGPT-2's shifts to layer 4 (L4MLP, L4H1, L5MLP) (Figure~\ref{fig:gt_dla_stacked}). DistilGPT-2's L4H1 replicates GPT-2's L9H1 comparison head with mean Pearson $r=0.945\pm0.079$ (median $r=0.981$), confirming the student embeds the teacher's comparison mechanism at the functionally analogous position in its shallower stack. Table~\ref{tab:gt_attention} reports per-example YY-token attention weights for five representative examples.

\begin{table*}[h]
\centering
\small
\caption{YY-token attention weights for DistilGPT-2 L4H1 and GPT-2 L9H1 at the prediction position, across five representative GT examples.}
\label{tab:gt_attention}
\begin{tabular}{lccccc}
\toprule
Year & YY & \multicolumn{2}{c}{Attention to YY token} & Pearson $r$ \\
\cmidrule(lr){3-4}
     &    & GPT-2 L9H1 & Distil L4H1 & (full pattern) \\
\midrule
1102 & 02 & 0.099 & 0.156 &- \\
1503 & 03 & 0.239 & 0.175 &- \\
1804 & 04 & 0.870 & 0.871 &- \\
1505 & 05 & 0.085 & 0.233 &- \\
1006 & 06 & 0.082 & 0.091 &- \\
\midrule
\multicolumn{4}{l}{Mean Pearson $r$ (full attention pattern, $n=1000$)} 
    & $0.945 \pm 0.079$ \\
\multicolumn{4}{l}{Median Pearson $r$} & $0.981$ \\
\multicolumn{4}{l}{Proportion with $r > 0.8$} & $93.6\%$ \\
\bottomrule
\end{tabular}
\end{table*}

Both heads consistently attend to the YY token across all five examples ($100\%$ detection rate), and the full-pattern Pearson correlation of $r = 0.945 \pm 0.079$ (median $0.981$) confirms that the student's L4H1 replicates the teacher's L9H1 at the level of individual attention weights, not merely in aggregate statistics.

\subsubsection{GT Circuit: Head-Level Activation Patching}
Year-encoding heads \{(0,1),(1,5)\} are the only GT circuit heads available in DistilGPT-2 (layers 0-5); all year-difference, comparison, and output-computation heads in layers 9-11 are absent, and no valid inter-head paths exist within layers 0-5, so path patching was not applicable.
\begin{table*}[h]
\centering
\small
\caption{GT circuit head-level activation patching results. Normalised scores
use noise ceiling $\hat{s} = s/({\rm GTProb}_{\rm clean} -
{\rm GTProb}_{\rm corrupt})$, noise ceiling $= 0.547$.}
\label{tab:gt_patching}
\begin{tabular}{llccccc}
\toprule
Head & Class & \multicolumn{2}{c}{Raw score} & 
              \multicolumn{2}{c}{Normalised score} & Sig. \\
\cmidrule(lr){3-4}\cmidrule(lr){5-6}
     &        & GPT-2 & Distil & GPT-2 & Distil & \\
\midrule
L0H1 & Year encoding & 0.332 & 0.283 & 0.606 & 0.517 & \checkmark \\
L1H5 & Year encoding & 0.305 & 0.255 & 0.557 & 0.465 & \checkmark \\
\midrule
\multicolumn{2}{l}{Overall (Cohen's $d$)} & \multicolumn{4}{c}{$d = 0.681$,\; $p < 0.0001$} & \checkmark \\
\bottomrule
\end{tabular}
\end{table*}

Both year-encoding heads show strong, significant causal contributions in DistilGPT-2 (normalised scores 0.517 and 0.465), comparable to GPT-2 (0.606 and 0.557). The overall effect size is large (Cohen's $d = 0.681$), confirming that the early-stage GT computation is causally preserved in the student. Path patching was not performed as no inter-head paths have both endpoints within layers 0-5.
\section{Discussion} \label{sec:discussion}
\subsection{RQ1: Representational Similarity After Depth Halving}
Three consistent patterns emerge across CKA, logit lens, attention analysis, and SAE feature statistics (\S\ref{sec:cka}--\S\ref{sec:sae}).

\textbf{Early layers are near-identical, late layers diverge.} Proportional-mapping CKA and per-head attention cosine similarity are both near 1.0 at the first layer pair, consistent with the student's every-other-layer initialisation, while same-index CKA collapses to 0.314 by layer 5 and SAE feature survival falls to 0.086 with reconstruction error $34.9\times$ the teacher's. The logit lens shows the same trend: same-index layer-matched representations become markedly less predictive of the next token with depth. 

\textbf{The student compresses rather than truncates.} The contrast between mapping schemes is itself informative: the same-index collapse at layer 5 is largely recovered under proportional mapping, showing that DistilGPT-2's final layer is more similar to GPT-2's layer 11 than to GPT-2's layer 5. The logit lens reinforces this proportional-mapping relative accuracy exceeds 1 at every layer but the last and functional SAE survival recovers to $\approx 0.53$ at mid- and late-pipeline pairs (magnitude drift $\delta \approx 0.51$--$0.55$). 

\subsection{RQ2: Circuit Preservation Under Depth Halving}
Distillation preserves circuits whose critical computations fit within the student's available depth and destroys those whose pipelines extend beyond it.

\textbf{The IOI circuit collapses completely.} The negative mean logit difference (\S\ref{sec:results_ioi}) is a sign reversal, not merely a performance gap, and DLA confirms no component provides meaningful positive attribution. Activation patching of the available heads yields negligible recovery, and name-mover head matching yields candidates with moderate CKA but no matched head reproduces IO-selective attention, despite some showing moderate-to-high attention-pattern correlation. The IOI circuit suffers from a mechanistic collapse: induction heads show a small, significant effect, but without downstream name-movers, the pipeline cannot complete.

\textbf{The GT circuit survives through active remapping.} In contrast, DLA reveals a coherent shift of peak contribution from the teacher's layer 9 to the student's layer 4, and the student's L4H1 replicates the attention pattern of GPT-2's L9H1 with mean Pearson $r = 0.945$, a level of alignment that stands in stark contrast to the IOI name-mover matching results. 

\textbf{Compressibility as a candidate explanation.} This dichotomy is consistent with a circuit compressibility account: circuits whose critical computations fit within the student's available depth survive, while those whose pipelines extend beyond it collapse. The result is structural incompleteness rather than graceful degradation. Confirming compressibility as a general principle requires testing intermediate-depth circuits across additional model pairs, which we identify as the most important extension of this work.

\section{Conclusion}
This work presented a mechanistic interpretability study of depth-reduction distillation, comparing GPT-2 and DistilGPT-2 across representational and causal probes. The finding is a compression-collapse dichotomy: distillation preserves circuits whose critical computations fit within the student's available depth and destroys those whose pipelines extend beyond it.
These findings suggest that aggregate metrics like perplexity are insufficient for assessing mechanistic preservation; practitioners should evaluate circuit integrity directly and use distillation objectives that align intermediate representations at functionally critical layers.

\section*{Limitations}

\paragraph{Scope.}
The analysis is restricted to two circuits and a single teacher-student pair. Whether the compression-collapse dichotomy generalises to other circuits, distillation methods, or model families remains open.

\paragraph{IOI null result.}
The IOI null result does not rule out DistilGPT-2 solving a residual version of the task via an alternative mechanism. Future work could apply causal scrubbing \citep{32} or automated circuit discovery \citep{20} to search for such subgraphs.

\paragraph{SAE reference basis.}
Using pre-trained GPT-2 SAEs as a fixed reference basis introduces a systematic blind spot to student-only features. Training student-specific SAEs would resolve this.

\paragraph{Incomplete verification of GT circuit survival.}
A complete mechanistic account of GT circuit survival would require verifying the full circuit beyond the key head match (L4H1 $\leftrightarrow$ L9H1). This includes the year-encoding heads $\{(0,1),(1,5)\}$ and the causal paths from year encoding through comparison to output computation, tested with both activation patching and path patching. Confirming that DistilGPT-2 replicates not just the comparison head's attention pattern but the full causal chain would substantially strengthen the claim of circuit survival.

\bibliography{custom}

\appendix
\section{Appendix}
\label{sec:appendix}

\subsection{Distillation Objective Interventions}
The analysis characterises the mechanistic consequences of soft-label distillation but does not intervene in the distillation process itself. An important open question is whether alternative objectives, intermediate-layer matching, attention transfer, or circuit-aware distillation losses could preserve circuit integrity during depth reduction. Preliminary evidence from the GT result suggests that even standard distillation can accidentally preserve circuits that are structurally compatible with the student's depth; targeted objectives might extend this to deeper circuits, such as IOI.

\subsection{SAE Streaming Metrics: Implementation Details}
All SAE-based statistics are computed in a single streaming pass over mini-batches of 64 activation vectors to avoid materialising the full sparse feature matrix $Z \in \mathbb{R}^{N_\text{tok} \times d_\text{SAE}}$ in memory. Feature survival (Pearson correlation) is computed via online sufficient statistics, running sums of $z_f$, $z_f^2$, and $z_f^\text{GPT-2} \cdot z_f^\text{distil}$, requiring only a single forward pass over the data. Activations are stored to disk after collection to avoid holding both model weights and SAE weights in GPU memory simultaneously; each SAE is loaded to the CPU and moved to the GPU only for the batch-level encode/decode operations. 

\subsection{CKA Full Results by Depth and Seed}

\begin{table*}
\centering
\small
\caption{CKA and cosine similarity by depth quartile under each mapping scheme, reported as mean across seeds \{42, 123, 7\}.}
\label{tab:cka_full}
\begin{tabular}{llcccc}
\toprule
Scheme & Depth & CKA & Cosine & CKA range & Cosine range \\
\midrule
Proportional & Early  & 0.9998 & 0.9977 &- &- \\
             & Middle & 0.9668 & 0.9749 &- &- \\
             & Late   & 0.7903 & 0.9690 &- &- \\
             & Overall & $0.921 \pm 0.109$ & $0.977 \pm 0.019$ 
                       & [0.662, 1.000] & [0.939, 0.999] \\
\midrule
Same-index   & Early  & 0.9998 & 0.9977 &- &- \\
             & Middle & 0.9082 & 0.9304 &- &- \\
             & Late   & 0.3138 & 0.3851 &- &- \\
             & Overall & $0.752 \pm 0.298$ & $0.784 \pm 0.310$ 
                       & [0.165, 1.000] & [0.057, 0.998] \\
\bottomrule
\end{tabular}
\end{table*}

Table~\ref{tab:cka_full} reports CKA and cosine similarity broken down by depth quartile under each mapping scheme.

\subsection{Attention Pattern Analysis: Numerical Summary}
Table~\ref{tab:attention_summary} provides numerical summaries of all attention pattern metrics.
\begin{table*}[h]
\centering
\small
\caption{Attention pattern metrics summary. Head similarity and entropy change reported as mean $\pm$ std across 1,000 TinyStories samples.}
\label{tab:attention_summary}
\begin{tabular}{lcc}
\toprule
Metric & GPT-2 & DistilGPT-2 \\
\midrule
Mean entropy (bits)      & 2.001 & 2.094 \\
Effective support (tokens) & 11.19 & 12.14 \\
Mean top-1 mass          & 0.507 & 0.493 \\
\midrule
\multicolumn{3}{l}{\textit{Cross-model head similarity (proportional mapping)}} \\
Mean cosine sim.\ & \multicolumn{2}{c}{$0.894 \pm 0.086$\ \ [0.610, 0.993]} \\
Mean $\Delta H$ (bits) & \multicolumn{2}{c}{$-0.007 \pm 0.621$} \\
\midrule
\multicolumn{3}{l}{\textit{Cross-model head similarity (same-index mapping)}} \\
Mean cosine sim.\ & \multicolumn{2}{c}{$0.631 \pm 0.255$\ \ [0.038, 0.981]} \\
Mean $\Delta H$ (bits) & \multicolumn{2}{c}{$-0.059 \pm 1.103$} \\
\bottomrule
\end{tabular}
\end{table*}
\subsection{SAE Feature Analysis Results}
\label{sec:results_sae1}

\begin{table*}[h]
\centering
\caption{SAE analysis summary across layer pairs.}
\label{tab:sae_summary}
\resizebox{\columnwidth}{!}{%
\begin{tabular}{llccc}
\toprule
Scheme & Pair & Cosine sim. & Feature survival & $\Delta L_0$ \\
\midrule
Same-index   & L0     & $0.989$ & $0.339$ & $-2{,}379$ \\
             & L3     & $0.887$ & $0.169$ & $-1{,}493$ \\
             & L5     & $0.655$ & $0.086$ & $-7{,}178$ \\
\midrule
Proportional & G1/D0  & $0.184$ & $0.000$ & $-132$ \\
             & G6/D3  & $0.927$ & $0.527$ & $-68$ \\
             & G11/D5 & $0.927$ & $0.529$ & $-601$ \\
\bottomrule
\end{tabular}%
}
\end{table*}
\end{document}